\documentclass{article}

\usepackage{arxiv}
\usepackage{cite}
\usepackage{amsmath,amssymb,amsfonts}
\usepackage{algorithmic}
\usepackage{graphicx}
\usepackage{textcomp}
\usepackage{xcolor}
\usepackage{tabularx}
\usepackage{hyperref} 
\usepackage[caption=false,font=footnotesize]{subfig}
\usepackage{booktabs} 
\usepackage{makecell} 
\usepackage[table]{xcolor}
\usepackage{multirow}
\usepackage{microtype}
\usepackage{siunitx}
\DeclareSIUnit{\degree}{\ensuremath{^\circ}}

\def\BibTeX{{\rm B\kern-.05em{\sc i\kern-.025em b}\kern-.08em
    T\kern-.1667em\lower.7ex\hbox{E}\kern-.125emX}}

\usepackage{xcolor}
\usepackage{graphicx}
\usepackage{booktabs}

\let\realincludegraphics\includegraphics

\newcommand{\modelname}{Apeliotes}
\newcommand{\degC}[1]{#1\ensuremath{^\circ}}
\newcommand{\nrmse}{\textsc{nrmse}}
\newcommand{\wspd}{\textsc{wspd}}
\newcommand{\mbe}{\textsc{mbe}}
\newcommand{\crps}{\textsc{crps}}
\newcommand{\mae}{\textsc{mae}}

\renewcommand{\includegraphics}[2][]{%
  \IfFileExists{#2}{%
    \realincludegraphics[#1]{#2}%
  }{%
    \par\medskip\noindent
    \fcolorbox{red}{yellow!20}{%
      \begin{minipage}{0.95\linewidth}
        \centering
        \small
        \textcolor{red}{\textbf{MISSING/DEPRECATED PLOT}}\par
        \texttt{\detokenize{#2}}\par
        \textit{Update the path.}
      \end{minipage}%
    }%
    \par\medskip
  }%
}

\usepackage[utf8]{inputenc} 
\usepackage[T1]{fontenc}    
\usepackage{hyperref}       
\usepackage{url}            
\usepackage{booktabs}       
\usepackage{amsfonts}       
\usepackage{nicefrac}       
\usepackage{microtype}      
\usepackage{lipsum}
\usepackage{graphicx}
\graphicspath{ {./images/} }

\title{Apeliotes: A Diffusion-Based Modeling Framework for km-scale Multi-Level Atmospheric Fields}

\author{
 Evangelia Rafaela Frastali, Achyut Paudel, Maryam Golbazi, Frank Liu \\
  Old Dominion University \\
  Norfolk, VA 23529, USA \\
  \texttt{\{efras001, apaudel, mgolbazi, fliu\}@odu.edu} \\
}

\begin{document}
\maketitle
\begin{abstract}
High-resolution atmospheric data are required to resolve mesoscale and localized meteorological structures, however such datasets remain limited in many regions of the world. Existing high-resolution weather products are typically produced through dynamical downscaling, which is computationally expensive and difficult to scale across locations, variables, and forecast scenarios. These limitations motivate machine-learning-based downscaling systems that can generate multiple weather variables stochastically while producing new high-resolution fields directly.
In this paper we present \modelname, a framework for high-resolution weather forecasting. Built on the global re-analysis atmospheric data, a pre-trained global weather foundation model, and a regionally trained generative diffusion model, \modelname~not only provides accurate kilometer-scale weather variables, but also multi-level atmospheric fields which are not directly available in the existing global atmospheric data. Our comprehensive evaluation demonstrates that \modelname~achieves highly competitive performance. The model predicts vertical wind profile with less than 3\% error between truth and predicted fields, achieving correlations of 0.91 for 10-m wind speed and 0.99 for 2-m temperature, with \nrmse~values of 0.42 and 0.17, respectively.
\end{abstract}

\keywords{Atmospheric Modeling \and Machine Learning \and Kilometer-Scale Forecasting \and Diffusion-based Downscaling \and Forecast Uncertainty}

\section{Introduction}
\label{sec:intro}
Weather modeling and forecasting support critical infrastructure across modern societies. Timely 
weather forecasting is essential to safeguarding public safety, enabling economic development, and enhancing regional resilience. Global weather modeling (at \degC{0.25}, \(\sim\)\qty{25}{\kilo\meter}, scale) is undergoing a fundamental transition enabled by recent advancements of machine learning (ML) and Artificial Intelligence (AI) solutions \cite{bodnar2025foundation, lam2023learning, price2025probabilistic}. However, kilometer-scale (km) weather forecasting is essential to resolve local variations in weather and climate, including the influence of land use, complex topography, and urban infrastructure, as well as interactions with population mobility and urban microclimate. Recently, a growing body of work has focused on downscaling coarse-resolution weather forecasts to kilometer-scale subgrid resolution \cite{park2026super,jha2025deep,gan2024machine,mardani2025residual}. Connecting global forecasts to regional weather is critical for delivering actionable information for local decision-making.

In this paper, we present \modelname, a framework which leverages a foundation model for efficient regional weather forecasting at \unit{km}-scale. \modelname~couples a global weather foundation model with a region-specific diffusion generative model that maps global, large-scale weather dynamics to \unit{km}-scale regional forecasts with substantially improved computational efficiency, enabling forecasts that traditionally require hours of computation to be produced in mere minutes. Moreover, the expressiveness of neural networks in the diffusion model enables direct data-driven learning of complex nonlinear multi-level atmospheric fields, which are not present in the global model outputs. 

The forecasting of multi-level fields at \unit{km}-scale opens the possibility of many impactful applications, ranging from aviation safety, wind energy forecasting and planning, weather prediction, air quality modeling, as well as boundary-layer dynamics. Trained with nine years of three-hourly reanalysis data, \modelname~demonstrated competitive skill, indicating that it can produce accurate high-resolution weather fields with performance comparable to established forecasting methods. Furthermore, \modelname~produces physically consistent forecasts for atmospheric fields, such as wind power density, thereby clearly highlighting the performance of our framework. The key contributions of this work are summarized as follows:

\begin{enumerate}
    \item We introduce \modelname, a framework for \unit{km}-scale regional weather forecasting, as well as its design and implementation.
    \item We extend a generative diffusion model to develop a regional modeling component that produces probabilistic kilometer-scale forecasts and reconstructs vertically resolved atmospheric fields beyond those explicitly provided by the global models.
    \item We demonstrate that the proposed framework learns physically meaningful cross-scale relationships, enabling large-scale atmospheric dynamics to multi-level atmospheric fields.
    \item We conduct a comprehensive skill evaluation showing that \modelname~is competitive with established baseline methods across variables.
\end{enumerate}
The rest of this paper is organized as follows: we discuss the architecture of \modelname~in Sec.~\ref{sec:details}, after a brief summary of existing work in Sec.~\ref{sec:background}. The paper is then followed by the training and implementation in Sec.~\ref{sec:setup} and experimental results in Sec.~\ref{sec:exp}. Finally, we discuss the limitations of our current work and future research in Sec.~\ref{sec:discussion}.

\section{Background and Related Work}
\label{sec:background}

Weather forecasting serves many practical applications, including aviation\cite{gultepe2019review}, environmental monitoring\cite{narayana2024advances}, pollution transport\cite{baklanov2020advances}, energy forecasting\cite{sweeney2020future}, and the analysis of extreme events\cite{sillmann2017understanding}, all of which require accurate high-resolution atmospheric data. Traditional numerical weather prediction (NWP)\cite{bauer2015quiet} uses physics-based models constrained by observations via data assimilation and refined with statistical post-processing~\cite{allen2025end, golbazi2024enhancing}. Despite its strong predictive performance, NWP is computationally expensive, which limits rapid forecasting, the use of large ensembles, and the development of task-specific high-resolution models. Furthermore, while NWP-based ensemble forecasts are feasible, they remain error-prone, require substantial engineering effort\cite{price2025probabilistic}, leaving substantial room for improvement. Data-driven approaches offer a complementary path; machine-learning models trained on NWP output and observations can generate or downscale forecasts at greatly reduced computational cost, enabling faster inference and broader downstream applications. Previous work shows that this strategy can closely approximate physical-model output across many environmental science tasks, including high-resolution weather modeling\cite{schmude2024prithvi,mardani2025residual, addison2022machine, hatanaka2023diffusion}. 

AI-based weather models can be grouped into three classes: deterministic predictive models, probabilistic generative models, and pretrained models that can be fine-tuned for specific tasks\cite{shi2025deep}. Our work combines the latter two. Foundation models for weather and climate with complex architectures are trained on heterogeneous Earth-system datasets and objectives, enabling successful outputs across variables, regions, and tasks. Representative examples include Aurora \cite{bodnar2025foundation}, Prithvi-WxC \cite{schmude2024prithvi}, ClimaX \cite{nguyen2023climax}, and GraphCast\cite{lam2023learning}.

Aurora is a 1.3-Billion-parameter weather foundation model partially trained on ERA5 data highlighted in \ref{subsec:Datasets}. It predicts key atmospheric variables with high fidelity and can generate 10-day global weather forecasts in minutes\cite{bodnar2025foundation}. We adopted Aurora as our large-scale foundation model because it has been broadly trained on atmosphere–ocean–land data and outperforms traditional methods on tasks such as air-quality forecasting, ocean-wave prediction, and tropical-cyclone track forecasting, while being substantially less computationally expensive than numerical solvers.
Aurora’s architecture consists of a Perceiver encoder, a 3D Swin Transformer backbone, and a Perceiver decoder. The encoder ingests heterogeneous inputs (multiple variables, spatial resolutions, pressure levels) and translates them into a 3D representation suitable as input for the 3D Swin Transformer. The 3D Swin Transformer produces an output, and the decoder translates it back to the native format of the input dataset. 

\begin{table}[tb]
\centering
\footnotesize
\setlength{\tabcolsep}{3pt}
\renewcommand{\arraystretch}{1.2}
\renewcommand\tabularxcolumn[1]{m{#1}}

\caption{Summary of selected relevant works. We report the downscaling models most closely related to the present study, along with their input and target grid spacings, high-resolution output pixel sizes, predicted variables, and model architectures.}

\label{tab:downscaling_models}
\begin{tabularx}{\columnwidth}{@{}
  >{\centering\arraybackslash}m{1.55cm}
  >{\centering\arraybackslash}m{1.75cm}
  >{\centering\arraybackslash}m{2.15cm}
  >{\centering\arraybackslash}X
@{}}
\toprule
\textbf{Reference} & \textbf{Architecture} & \textbf{Input $\rightarrow$ Target} & \textbf{Variables} \\
\midrule
Mardani et al.~\cite{mardani2025residual}
& Residual diffusion
& 25 $\rightarrow$ 2 km ($448{\times}448$)
& 10-m wind ($u, v$), 2-m temperature, Radar reflectivity \\
\addlinespace
Lopez-Gomez et al.~\cite{lopez2025dynamical}
& Dynamical + diffusion
& 45 $\rightarrow$ 9 km ($340{\times}270$)
& 2-m temperature, 2-m specific humidity, 10-m winds ($u, v$), Surface pressure, Precipitation \\
\addlinespace
Schmude et al.~\cite{schmude2024prithvi}
& Transformer (ViT)
& 300 $\rightarrow$ 50 km ($360{\times}576$)
& 2-m temperature \\
\addlinespace
Aich et al.~\cite{aich2026conditional}
& Conditional diffusion
& 100 $\rightarrow$ 25 km ($256{\times}256$)
& Precipitation \\
\midrule
\textbf{Our work}
& Generative diffusion
& 25 $\rightarrow$ 4 km ($112{\times}112$)
& 22 multi-level atmospheric fields \\
\bottomrule
\end{tabularx}
\end{table}

Coarse resolution climate models (numerical and AI) operating at $\ge$\qty{25}{\kilo\meter} scales often fail to resolve \unit{km}-scale phenomena and local trends. While high-resolution data are crucial for regional forecasting, larger ensemble sizes, improved climate downscaling, and creating detailed forecasts in data-scarce areas\cite{mardani2025residual}, generating them via regional NWP is computationally prohibitive over long periods\cite{schmude2024prithvi}. AI models offer a more efficient alternative. However, they often rely on high-quality reanalysis data that may be unavailable in data-scarce regions. To address these limitations, downscaling models have been developed to map coarse inputs to fine-scale outputs\cite{giorgi2019thirty} that capture regional-level phenomena. These approaches have evolved from convolutional architectures to generative methods, such as diffusion models\cite{lopez2025dynamical,nguyen2023climax,mardani2025residual,tu2025satellite}. Tab.~\ref{tab:downscaling_models}
summarizes weather downscaling models closely related to the proposed method. Building on these advancements, our goal is to produce high-resolution, multi-channel output that overcomes these computational and data constraints.

\section{Details of Apeliotes}
\label{sec:details}

\subsection{\modelname: a Framework for Regional Forecasting}

\begin{figure}[htbp]
\centerline{\includegraphics[width=0.99\linewidth]{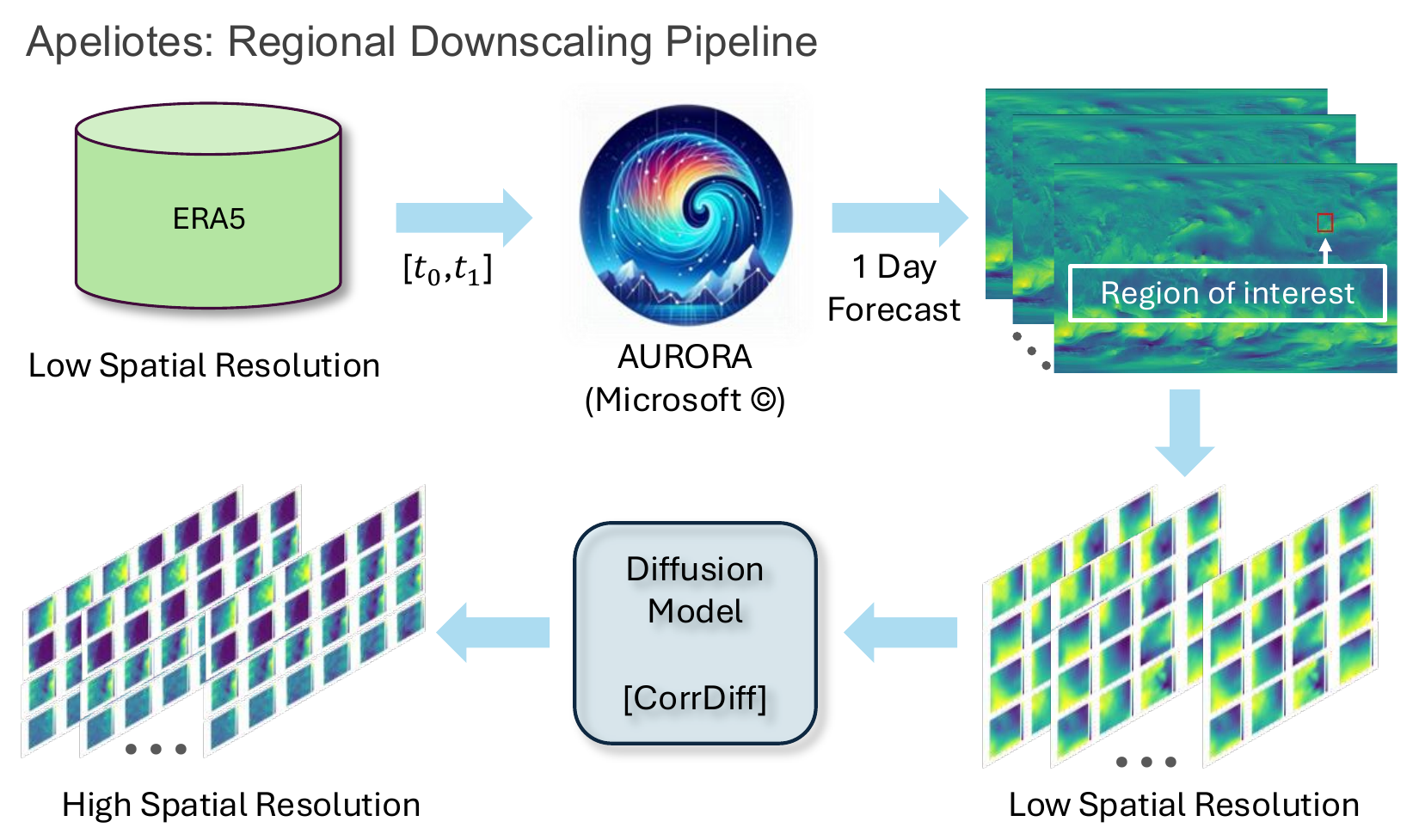}}
\caption{Architecture of \modelname. The Aurora model ingests ERA5 data to produce coarse-resolution global forecasts ($\sim$\qty{25}{\kilo\meter}), which are cropped to the domain of interest and downscaled via CorrDiff to probabilistic \qty{4}{\kilo\meter} regional outputs. Using an autoregressive approach, Aurora advances two timesteps of coarse-resolution ERA5 data to generate a one-day forecast, which is downscaled to \qty{24}{\hour} of regional high-resolution data at 6-hour intervals.}
\label{fig:pipeline}
\end{figure}

In this work, we present a novel framework that combines a foundation model for the Earth system (Aurora at $\sim$\qty{25}{\kilo\meter} resolution) with the Corrective Diffusion (CorrDiff) downscaling model to enable short-timescale regional weather predictions at 4\,km resolution for the Eastern US. The overall architecture of \modelname~ is shown in Fig.~\ref{fig:pipeline}. We use the pretrained Aurora model in inference-only mode, with ERA5 as input for autoregression. We trained CorrDiff with a custom pipeline using nine years of paired ERA5-WRF data in ~\ref{subsec:Datasets} for a region covering the eastern U.S, mostly the Hampton Roads domain.

\subsection{Datasets and Models}
\label{subsec:Datasets}
\noindent
\textit{ERA5 reanalyses dataset:} To train the diffusion model, we use ERA5 dataset as the input
data. ERA5 is the fifth-generation global atmospheric reanalysis product of the European Center
for Medium-Range Weather Forecasts (ECMWF), providing hourly atmospheric and land-surface
variables from 1940 to present at \degC{0.25} horizontal resolution with uncertainty estimate
that is sampled by an underlying 10-member ensemble~\cite{hersbach2023era5}.
By combining observations with NWP model forecasts through data assimilation, ERA5 is a reanalysis dataset widely accepted as a leading global reconstruction of the historical atmospheric state.\\
\textit{WRF dataset}: For the target dataset during the training, we use the long-term \qty{4}{\kilo\meter} resolution hydro-climate reanalyses generated with WRF and obtained from the National Center for Atmospheric Research Geoscience Data Exchange (NCAR-GDEX)~\cite{ncar_gdex_dataset_d612005}. The dataset is produced using ERA5 reanalysis fields as initial and boundary conditions in WRF and provides physically consistent atmospheric and land-surface variables at convection-permitting scales suitable for mesoscale meteorological analyses. The simulation spans multiple decades and includes hourly three-dimensional atmospheric variables, surface fields, and static geographic information stored in NetCDF format on a Lambert Conformal grid covering the contiguous United States (CONUS) and surrounding regions. \\
\textit{Aurora Foundation Model:} 
\modelname~ uses Aurora’s $\sim$\qtyrange{25}{31}{\kilo\meter} resolution forecasts as real-time inputs to produce fine-scale predictions for the Eastern U.S. within minutes, avoiding the high computational cost of generating new ERA5 analyses. In this setup, ERA5 acts as the physically grounded training reference, while Aurora provides the input for real-time downscaling. This approach preserves the physical consistency of model output in the learned mapping while enabling speed and scalability.

\subsection{Downscaling from global model to regional model}

Downscaling from global to regional models has traditionally relied on dynamical downscaling, a computationally expensive approach. As an alternative, statistical downscaling leverages simulations from dynamical models and observational data to learn the parameters of a statistical mapping\cite{mardani2025residual}. ML models offer a powerful extension to this approach, as they can capture nonlinear relationships and thus outperform classical statistical methods. Moreover, generative ML models are particularly well-suited for representing the stochastic nature of atmospheric physics at km-scale resolutions\cite{mardani2025residual}.

In \modelname, we train the CorrDiff model from scratch for the target region. CorrDiff is a residual corrective diffusion model for km-scale atmospheric downscaling. In CorrDiff, first a deterministic U-Net regression model is used to predict the coarse-to-fine mapping (the mean downscaled field), then a stochastic diffusion model is applied to add realistic fine-scale structure as residual correction\cite{mardani2025residual}. The U-Net model captures the large-scale mean component while the diffusion step generates small-scale perturbations consistent with learned physics, thereby producing ensemble-consistent local details that are absent at $\sim25$\,km resolution. 

\modelname~leverages a custom machine learning pipeline and extensive GPU training to surpass classical approaches. Traditional statistical downscaling methods typically generate high-resolution outputs for only a small subset of variables from a larger low-resolution set. As analyzed by\cite{mardani2025residual}, most existing models downscale to one or two output channels, with CorrDiff representing the state of the art by producing four. In contrast, our trained downscaling model extends this capability significantly, generating 22 high-resolution output channels from only 17 low-resolution input channels, see Table~\ref{tab:wpd-channels} for details.

This demonstrates the ability of our model, a generative statistical downscaler combined with Aurora, to create a complete high-resolution state vector at time $t +\delta t$ using both high and low-resolution states at time $t$. Importantly, this capability, previously achievable only through computationally intensive dynamical downscaling over hundreds of autoregressive steps, can now be accomplished in a single inference.
In \modelname, we further extend the capability of CorrDiff-type downscaler by exploring \emph{vertical} downscaling. Many global models provide wind forecasts at a coarse vertical resolution (e.g., ERA5 has values at \qty{925}{\milli\bar}, \qty{1000}{\milli\bar}, etc.), which is insufficient for applications that require detailed vertical wind profiles close to the surface. For instance, this resolution is insufficient for resolving wind speed profiles within the critical wind turbine rotor layer (up to $\sim$\qty{230}{\meter}), where detailed vertical layers are essential for wind energy assessment. In \modelname, we train CorrDiff to generate high-resolution vertical wind profiles from coarse profile inputs. Specifically, using the 3D output of WRF, we construct vertical slices (e.g., 7 vertical levels within the first layer of ERA5). We further generalize this capability to other atmospheric fields, as outlined in the next subsection.

\subsection{Modeling of Multi-level Atmospheric Fields}

\begin{table}[tb]
\centering
\footnotesize
\renewcommand{\arraystretch}{1.15}
\caption{Input (ERA5) and output (WRF) variables, grouped by single-level and pressure-level fields. 17 input channels and 22 ouput channels on a shared $112 \times 112$ grids. }
\label{tab:wpd-channels}
\begin{tabular*}{\columnwidth}{@{\hspace{4pt}}l@{\extracolsep{\fill}}lcc@{\hspace{4pt}}}
\toprule
\textbf{Description} & \textbf{Units} & \textbf{Input (ERA5)} & \textbf{Output (WRF)} \\
\midrule
Pixel resolution       & ---                       & $112 \times 112$               & $112 \times 112$ \\
\midrule
\multicolumn{4}{@{\hspace{4pt}}l}{\emph{Single-level fields}} \\
2\,m temperature       & K                         & \texttt{t2m} & \texttt{T2} \\
10\,m eastward wind    & m\,s$^{-1}$               & \texttt{u10} & \texttt{U10} \\
10\,m northward wind   & m\,s$^{-1}$               & \texttt{v10} & \texttt{V10} \\
10\,m wind speed       & m\,s$^{-1}$               & ---          & \texttt{SPEED10} \\
\midrule
\multicolumn{4}{@{\hspace{4pt}}l}{\emph{Pressure-level fields}} \\
Temperature            & K                         & \texttt{t} at levels 0,\,1,\,2 & \texttt{TK} at levels 0,\,1 \\
Eastward wind          & m\,s$^{-1}$               & \texttt{u} at levels 0,\,1,\,2 & \texttt{U} at levels 0--6 \\
Northward wind         & m\,s$^{-1}$               & \texttt{v} at levels 0,\,1,\,2 & \texttt{V} at levels 0--6 \\
Specific humidity      & kg\,kg$^{-1}$             & \texttt{q} at levels 0,\,1,\,2 & --- \\
Geopotential height    & m$^{2}$\,s$^{-2}$         & \texttt{Z} at levels 1,\,2     & --- \\
Wind power density     & W\,m$^{-2}$               & ---                            & \texttt{WPD} at levels 1,\,2 \\
\midrule
\textbf{Total channels} &                       & \textbf{17}                    & \textbf{22} \\
\bottomrule
\end{tabular*}
\end{table}

In \modelname, we extend the generative diffusion–based downscaler to produce complex atmospheric fields in addition to standard prognostic variables. These fields are typically treated as diagnostic outputs in NWP models, and are often nonlienar function of other outputs. By directly learning and generating these fields alongside the three‑dimensional state variables, the model provides physically consistent, application‑relevant outputs at high spatial resolution. In \modelname, all multi‑level fields are generated directly at \qty{4}{\kilo\meter} resolution, including the three fields described below. The direct generation of these fields is the distinctive strength of \modelname~and extend its capabilities beyond a mere downscaling framework.

\subsubsection{Wind Power Density}

Wind power density (WPD) is a commonly used metric in wind resource assessment that represents the kinetic energy available in the wind per unit area \cite{zhang2014study, dabbaghiyan2016evaluation, shu2015investigation}. Compared to wind speed alone, WPD provides a more direct measure of wind energy potential because it accounts for the cubic dependence of power on wind speed. Wind power density is defined as:

\begin{equation}
\mathrm{WPD} = \frac{1}{2}\,\rho v^3 ,\label{eq1}
\end{equation}
where $v$ is wind speed in \unit{\meter\per\second} and $\rho$ is air density in \unit{\kilogram\per\cubic\meter}. The units of WPD are \unit{\watt\per\square\meter}, which is the amount of wind energy available per unit area perpendicular to the airflow.

The calculation of WPD requires an estimate of air density, which in turn depends on air pressure, temperature, and mixing ratio. We use the following equation to estimate the air density at every grid cell \cite{hobbs2006atmospheric}:
\begin{equation}
\rho = \frac{\epsilon p(1+w)}{R_d T(w + \epsilon)}\label{eq2}
\end{equation}

where $p$ is air pressure in \unit{\pascal}, $R_d$ is the specific gas constant for dry air, $T$ is air temperature in K, $w$ is the mixing ratio, and ($\epsilon \approx 0.622 $) is the molecular weight ratio. Here the WPD calculation is performed consistently with the atmospheric state variables used during training to maintain physical consistency.

\subsubsection{Wind Speed}
Wind speed is a nonlinear function of the zonal ($u$) and meridional ($v$) components of the three-dimensional wind field calculated as $ws = \sqrt{u^2+v^2}$, as described in Tab.~\ref{tab:wpd-channels}. In \modelname, we predict not only the wind components at multiple vertical levels, but also directly generate wind speed at the 10-m level, which is commonly used for surface wind assessments. While this work focuses on 10-m wind speed, the model architecture can be easily adapted to predict wind speed at other vertical levels, depending on downstream needs or application-specific requirements.

\subsubsection{Vertical Wind Profile}

\modelname\ also captures the $u$ and $v$ wind components to reconstruct vertical wind profiles at seven pressure levels within the lower atmosphere, which fits within the first vertical level of the input model. By learning the vertical structure of the wind field from the high‑resolution reference data, the model is able to infer additional vertical detail beyond what is explicitly available in the input. This capability enables a more complete representation of near-surface and lower-tropospheric wind profiles, which is critical for applications such as wind resource assessment \cite{golbazi2022surface} and boundary-layer characterization\cite{golbazi2019methods}. 
In addition, a performance analysis of our model to predict the vertical profiles is illustrated in Fig.~\ref{fig:profile_combined} with ERA5 and AURORA as inputs, respectively.

\section{Experimental Setup}
\label{sec:setup}

We build on Aurora, a recently introduced large-scale foundation model for Earth system forecasting, but train our downscaling system using ERA5 reanalysis rather than Aurora’s own forecasts. ERA5, produced by ECMWF, provides hourly global atmospheric fields at  \degC{0.25} ($\sim$\qtyrange{25}{31}{\kilo\meter}) resolution by combining a numerical weather prediction model with a wide range of historical observations.
As high-resolution target data for training, we use the NCAR RDA d612005 dataset \cite{rasmussen2021conus}, a mesoscale hydro-climate simulation over the CONUS, generated by dynamically downscaling CMIP5 ensemble mean boundary conditions using the WRF model at 4\,km horizontal resolution. 
Spatial coverage spans the CONUS domain using a Lambert Conformal projection ($ 1429\times1419$ grid cells and 50 vertical levels) with \qty{4}{\kilo\meter} horizontal grid resolution. Using simulation data as target data is necessary because observations, including satellite and in situ sensor measurements, are spatially and temporally sparse at high-resolution grid scales and are therefore not sufficient on their own to train and test models. In this work, we will use a part of the WRF dataset over our domain of interest, which contains the Northeast United States. The present domain includes mountainous areas, urban regions, and both coastal and inland terrain, making it topographically diverse. Aurora outputs are on the  \degC{0.25} grid (matching ERA5), ensuring resolution compatibility with our training data, while WRF provides higher-resolution fields suited for local atmospheric applications. This setup enables Apeliotes to learn km-scale structure while maintaining sufficient fidelity for our target use cases.

To enable CorrDiff to ingest Aurora model output, we construct a CorrDiff-compatible Zarr dataset from Aurora NetCDF files. The objective is to reproduce the structure, coordinates, and normalization conventions of the CorrDiff training dataset while replacing ERA5-based inputs with Aurora predictions.

Dataset Description: The CorrDiff training Zarr contains (i) static fields, (ii) normalization parameters (e.g., era5\_center, era5\_scale, wrf\_center, wrf\_scale), and (iii) geolocation fields such as latitude (XLAT) and longitude (XLONG). In our pipeline, these components are reused and copied onto the Aurora outputs. This ensures that the resulting Aurora Zarr shares the same coordinate system, scaling factors, and metadata layout as the original training data.

Aurora predictions are provided on Aurora’s native latitude–longitude grid, which generally differs from the WRF grid used by CorrDiff. We employ ESMF-based bilinear interpolation (xESMF) to map Aurora fields onto the CorrDiff WRF domain.

To handle large-scale, multi-resolution datasets used for training and evaluation, we adopt a simple yet effective data management strategy that standardizes preprocessing, versioning, and I/O. All input, output, and target datasets are stored in local storage within the DGX H100 node, so training and inference run entirely on a single node, eliminating network latency, file transfer overhead, and cross-machine scheduling or partitioning. For subsequent validation and reproducibility experiments, we replicated this setup on the NVIDIA Cloud Brev platform with A100 GPUs, allowing us to rerun and extend our experiments under the same data-management scheme.

The training pipeline expects paired coarse and high-resolution data packaged as a Zarr store. Our training Zarr store contains 3-hourly samples from 2007–2015, nine years in total, with 26,996 timestamps, reduced to 25,513 after removing entries without a valid coarse- and high-resolution pair. We use 2007–2013 for training and 2014–2015 for validation/testing. 
CorrDiff is comprised of two components: a regression network and a diffusion network that conditions on the regression output. Both are U-Net architectures with ($\sim$80M parameters). We retain\cite{mardani2025residual} optimizer and schedule (Adam + warm-up) but train using a custom pipeline. All configuration, training, and testing were run on a DGX node with 8$\times$H100 GPUs (batch size 512); inference uses a single GPU.

\section{Experimental Results}
\label{sec:exp}

We trained the generative component of \modelname~ from scratch. %
The details of the trained model, with 17 input and 22 output channels, are listed in Tab.~\ref{tab:wpd-channels}. Together with the global model Aurora, \modelname~is capable of providing near surface atmospheric variables, including 2-m temperature (surface temperature), $u$ and $v$ wind components at 10-m, at a fraction of the traditional NWP method. More importantly, \modelname~is also capable of generating other atmospheric fields which are not present in the global model, including surface wind speed, wind components at different levels, and WPD at two levels. 

\subsection{Model Evaluation}

The evaluation of \modelname~ was completed in two stages: the custom-trained generative downscaler and the overall model performance. 
To assess the performance and generalization capability of the trained CorrDiff-based downscaling model, we conducted an out-of-sample evaluation using data from the years 2014 and 2015. These years were intentionally excluded from the training period to provide an independent validation dataset. The custom-trained CorrDiff diffusion model was applied to the coarse-resolution \degC{0.25} ERA5 fields for these two years to generate downscaled outputs for all output channels, including the key meteorological variables considered in this study, i.e., the $u$ and $v$ wind components at seven vertical levels, water vapor mixing ratio, and 2-m air temperature. 

In atmospheric science, skill analysis is widely used as a quantitative assessment of how accurately a model reproduces atmospheric states or predicts their evolution. It compares model outputs with observations or trusted reference datasets and measures improvements over baselines. For model skill analysis, we select wind speed and wind direction diagnosed from the $u$ and $v$ components as primary variables for detailed evaluation due to their complexity to forecast and their relevance to the boundary-layer processes, the high level task of predicting wind power density, and regional wind climatology. For spatial model performance analysis, we will focus on 2-m temperature, 10-m wind speed, and vertical wind profile prediction, due to their importance and relevance to the other fields.

Because the objective of this work is to reproduce high-resolution (\qty{4}{\kilo\meter}) atmospheric fields consistent with those produced by the WRF model, the WRF simulations are treated as the reference or "ground truth" for model evaluation. The goal is not to quantify or correct biases inherent to either ERA5 or WRF, but to assess how closely the trained downscaling component can approximate 4-km WRF fields when provided with \degC{0.25} ERA5 or Aurora inputs. Comprehensive evaluations of WRF model performance for this region has been thoroughly done and available in the literature \cite{carvalho2012sensitivity, bhati2016wrf, golbazi2022surface, golbazi2024large, golbazi2025high} and are therefore beyond the scope of this study. 

\begin{figure}[htpb]
    \centering
    \subfloat[ERA5]{\includegraphics[width=0.99\linewidth]{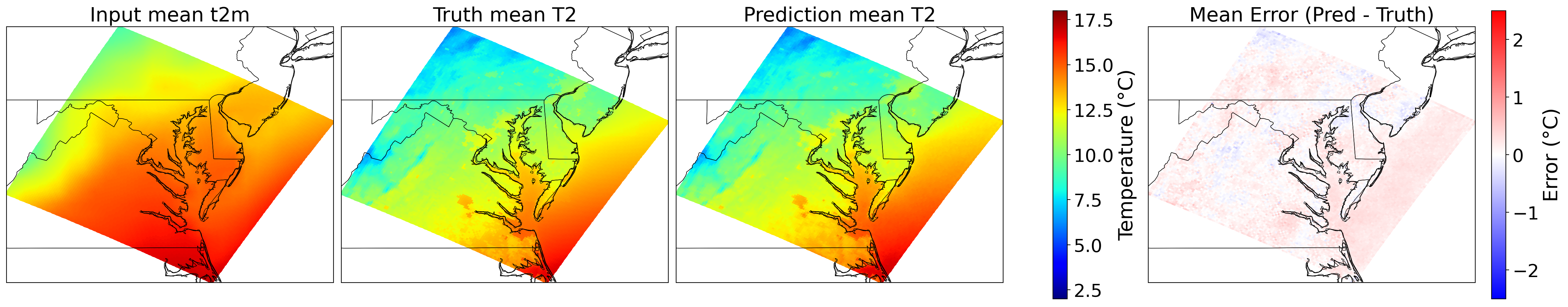}\label{fig:2m_temperature_era5}}

    \subfloat[Aurora]{\includegraphics[width=0.99\linewidth]{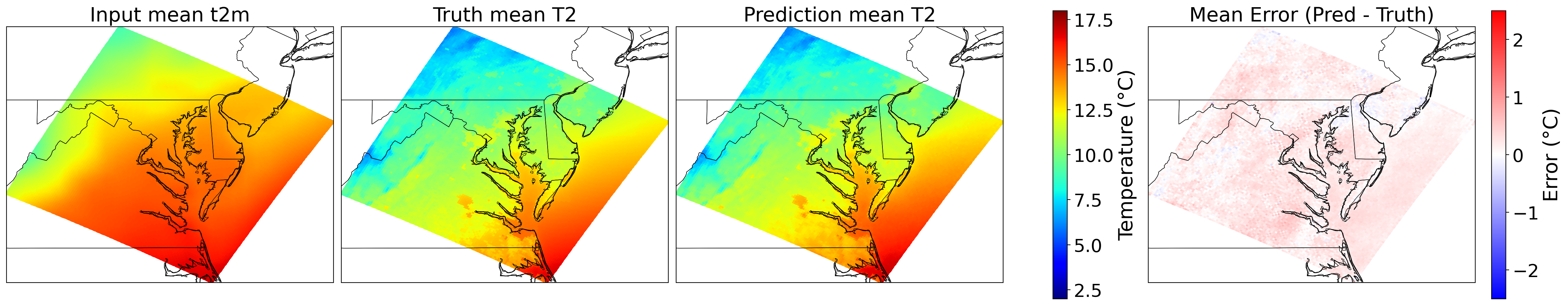}\label{fig:2m_temperature_aurora}}
    
    \caption{Input, reference, and predicted 2-m temperature (deg~C) fields using ERA5 (top row) and Aurora (bottom row) as input data. For each row, the panels show the input, reference, prediction, and error (prediction $-$ truth), from left to right, respectively. The predictions capture fine-scale spatial variability across the domain, including enhanced temperatures over urban areas.}
    \label{fig:2m_temperature}
\end{figure}

\begin{figure}[htp]
    \centering
    \subfloat[ERA5]{\includegraphics[width=0.99\linewidth]{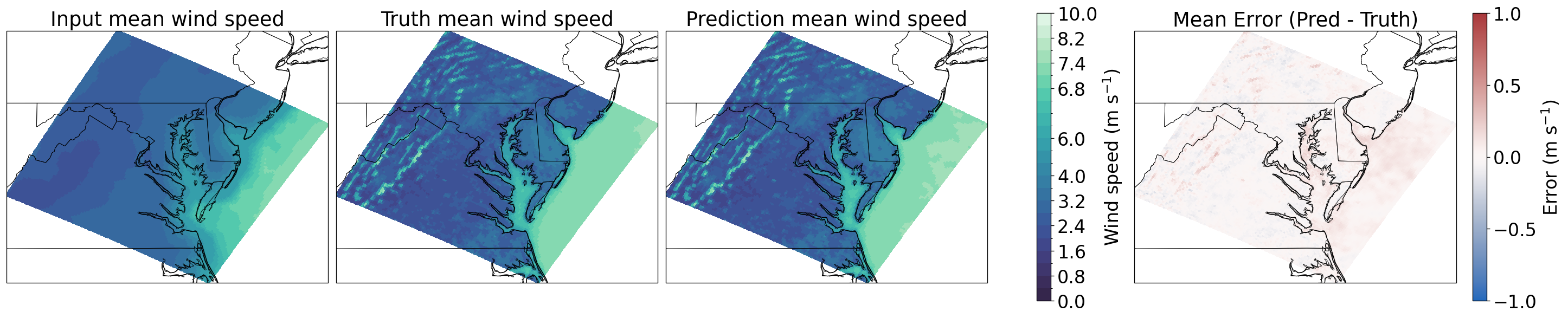}\label{fig:10m_wind_era5}}

    \subfloat[Aurora]{\includegraphics[width=0.99\linewidth]{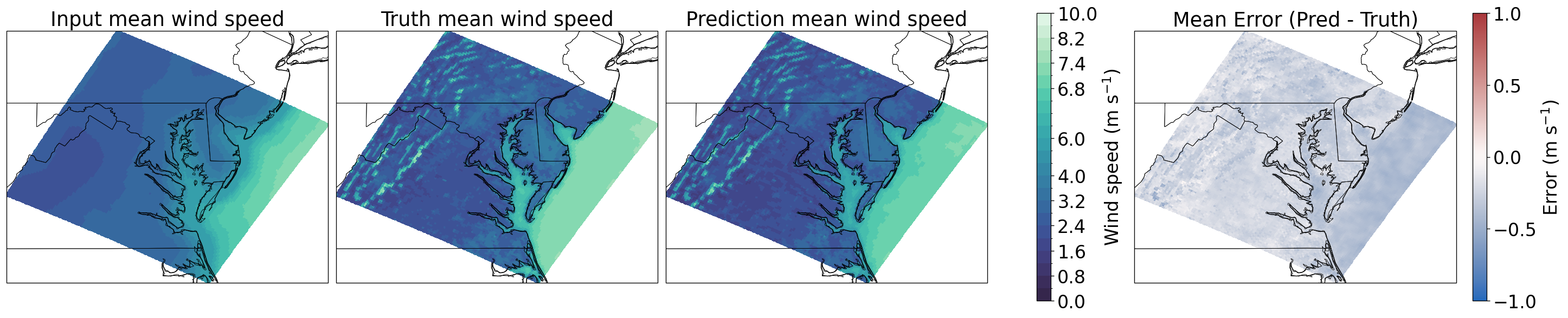}\label{fig:10m_wind_aurora}}
    
    \caption{Input, reference, and predicted 10-m wind speed fields using ERA5 (top row) and Aurora (bottom row) as input data. For each row, the left, middle, and right panels show the input, reference, and predicted wind speed, respectively.}
    \label{fig:10m_wind}
\end{figure}

After training, we evaluated the robustness of the downscaling framework by generating downscaled outputs from two different input sources: ERA5 and Aurora. ERA5 and Aurora inference were applied across the full validation period to study the performance of the model. The results for 2-m temperature and 10-m wind speed are illustrated in Fig.~\ref{fig:2m_temperature} and \ref{fig:10m_wind}, respectively. As a result, the spatial patterns in the Aurora-driven outputs (bottom row in Fig. \ref{fig:2m_temperature} and \ref{fig:10m_wind}) have some differences relative to the ERA5-driven results (top row in Fig. \ref{fig:2m_temperature} and \ref{fig:10m_wind}). Despite the uncertainties and biases inherent to Aurora-based runs, downscaled fields from both inputs show strong agreement with the WRF target, indicating that the model consistently captures fine-scale spatial variability. In particular, the downscaled temperature fields reproduce coherent thermal structures and enhanced heat concentrations over urban areas, indicating that the model effectively recovers localized surface features and mesoscale detail regardless of the input data source. In addition, the downscaled wind speed fields capture fine-scale flow structures and topography-driven variations, indicating that the model effectively reconstructs terrain-induced wind patterns and mesoscale detail, regardless of the input data source. 

\subsection{Skill Analysis}

The CorrDiff model can generate ensembles of any chosen size; we used a 32-member ensemble for each output channel because the original CorrDiff paper reported limited gains beyond this size. Unless stated otherwise, we use the ensemble mean for the evaluation, as it provides a stable and representative estimate of the downscaled field.

\begin{table}[!b]
\centering
\caption{Normalized deterministic and probabilistic metrics for 10-m wind speed (\wspd) and 2-m temperature. Lower \nrmse, \mbe, and \crps~indicate better performance, while higher correlation indicates stronger agreement with the reference fields.}
\label{tab:metrics}
\footnotesize
\setlength{\tabcolsep}{1.5pt}
\renewcommand{\arraystretch}{1.2}
\begin{tabularx}{\columnwidth}{c *{8}{>{\centering\arraybackslash}X}}
\toprule
\multirow{2}{*}{\textbf{Input}}
& \multicolumn{4}{c}{\textbf{10-m WSPD [m\,s$^{-1}$]}} 
& \multicolumn{4}{c}{\textbf{2-m T [$^\circ$C]}} \\
\cmidrule(lr){2-5} \cmidrule(lr){6-9}
& NRMSE & MBE & Corr & CRPS
& NRMSE & MBE & Corr & CRPS \\
\midrule
ERA5
& 0.25 & 0.01 & 0.97 & 0.37
& 0.11 & 0.11 & 0.99 & 0.68 \\
Aurora
& 0.42 & $-0.27$ & 0.91 & 0.66
& 0.17 & 0.18 & 0.99 & 1.23 \\
\bottomrule
\end{tabularx}
\end{table}

Table~\ref{tab:metrics} summarizes deterministic and probabilistic performance metrics for two selected variables, 10-m wind speed and 2-m temperature predictions, for two inference configurations using ERA5 and Aurora inputs. The normalized root-mean-square error (\nrmse) evaluates deterministic prediction accuracy relative to the underlying variability of the reference field and is computed as 
\begin{equation}
    \nrmse = \frac{\sqrt{\frac{1}{N}\sum_{i=1}^{N}(y_i-\hat{y}_i)^2}}{\sigma_y},
    \label{eq:nrmse}
\end{equation}

where $y_i$ and $\hat{y}_i$ denote the reference and predicted values, respectively, and $\sigma_y$ is the standard deviation of the reference field. Mean bias error (\mbe) quantifies systematic over- or under-prediction, while $Corr$ represents the Pearson correlation coefficient between predictions and reference values. Probabilistic skill is evaluated using the Continuous Ranked Probability Score (\crps) \cite{alessandrini2015analog} given as 

\begin{equation}
\crps=\frac{1}{N}\sum_{i=1}^{N}\int_{-\infty}^{\infty}[F_i^f(x)-F_i^0(x)]^2\,dx,
\label{eq:crps}
\end{equation}

where $F_i^f(x)$ is the forecast cumulative distribution function (CDF), $F_i^0(x)$ is the CDF of the observation for the $i^{th}$ ensemble prediction/observation pair, and $N$ is the number of available pairs. It is shown in the literature how the \crps~reduces to the mean absolute error (\mae) for a deterministic forecast~\cite{hersbach2000decomposition}. A Lower \crps~value indicates better performance.


To assess the model’s ability to reproduce local wind speed and direction, we generated polar histograms for both the WRF wind data and the corresponding predictions. The polar histogram represents the frequency of wind occurrence as a function of direction, where each angular bin corresponds to a directional sector and the radial distance indicates the relative frequency of winds from that sector. Comparison of the input and predicted distributions is used to assess whether the model preserves the dominant wind directions and their frequency of occurrence. As illustrated in Fig.~\ref{fig:profile_combined}c and Fig.~\ref{fig:profile_combined}f, the predictions closely align with the reference, indicating that the model accurately captures the dominant wind direction 

\subsection{Prediction of Multi-Level Atmospheric Fields}
\begin{figure}[t]
    \centering

    \subfloat[]{%
        \includegraphics[
            width=0.31\linewidth,
            trim=20 10 20 10,
            clip
        ]{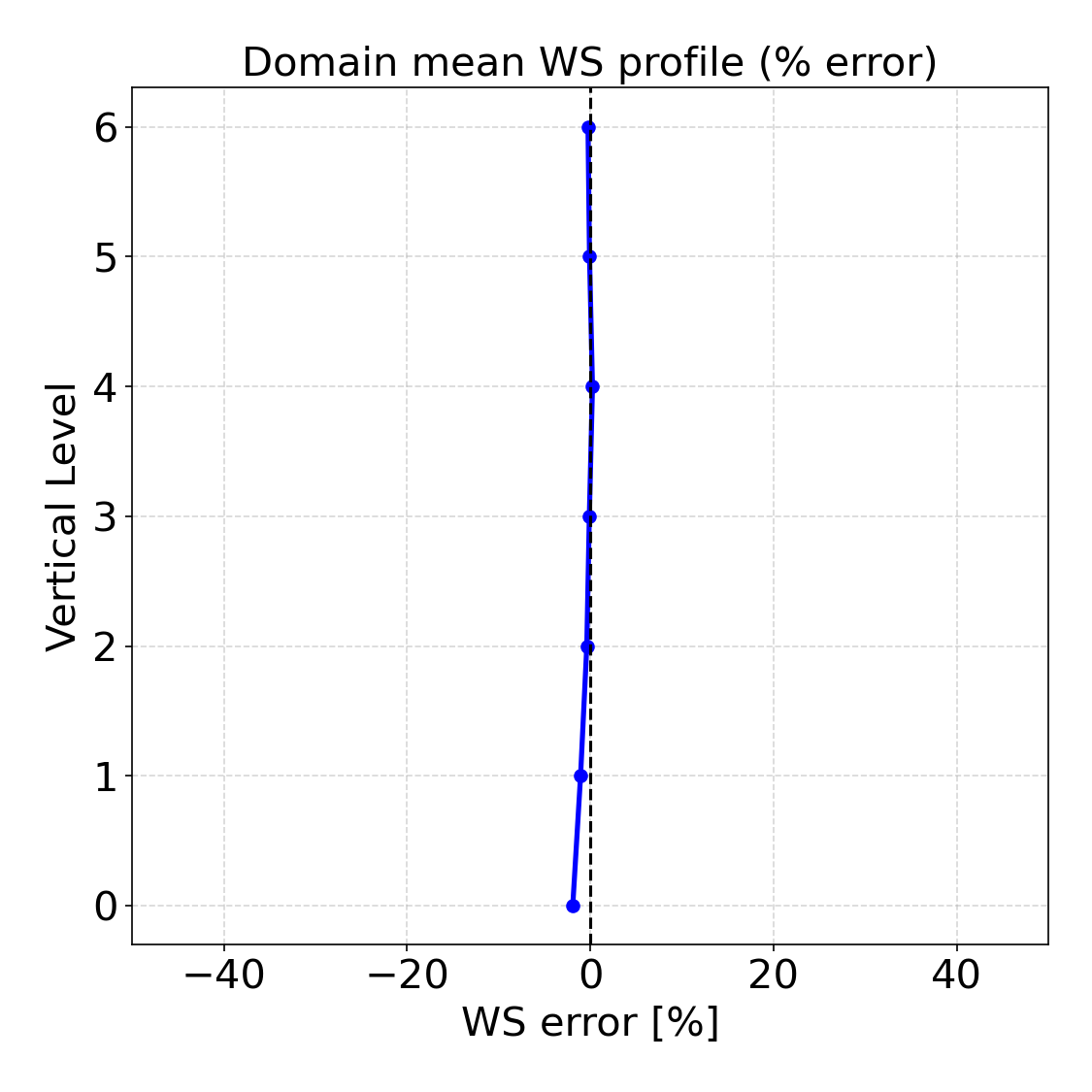}
        \label{fig:era5_profile_WS}
    }\hfill
    \subfloat[]{%
        \includegraphics[
            width=0.31\linewidth,
            trim=20 10 20 10,
            clip
        ]{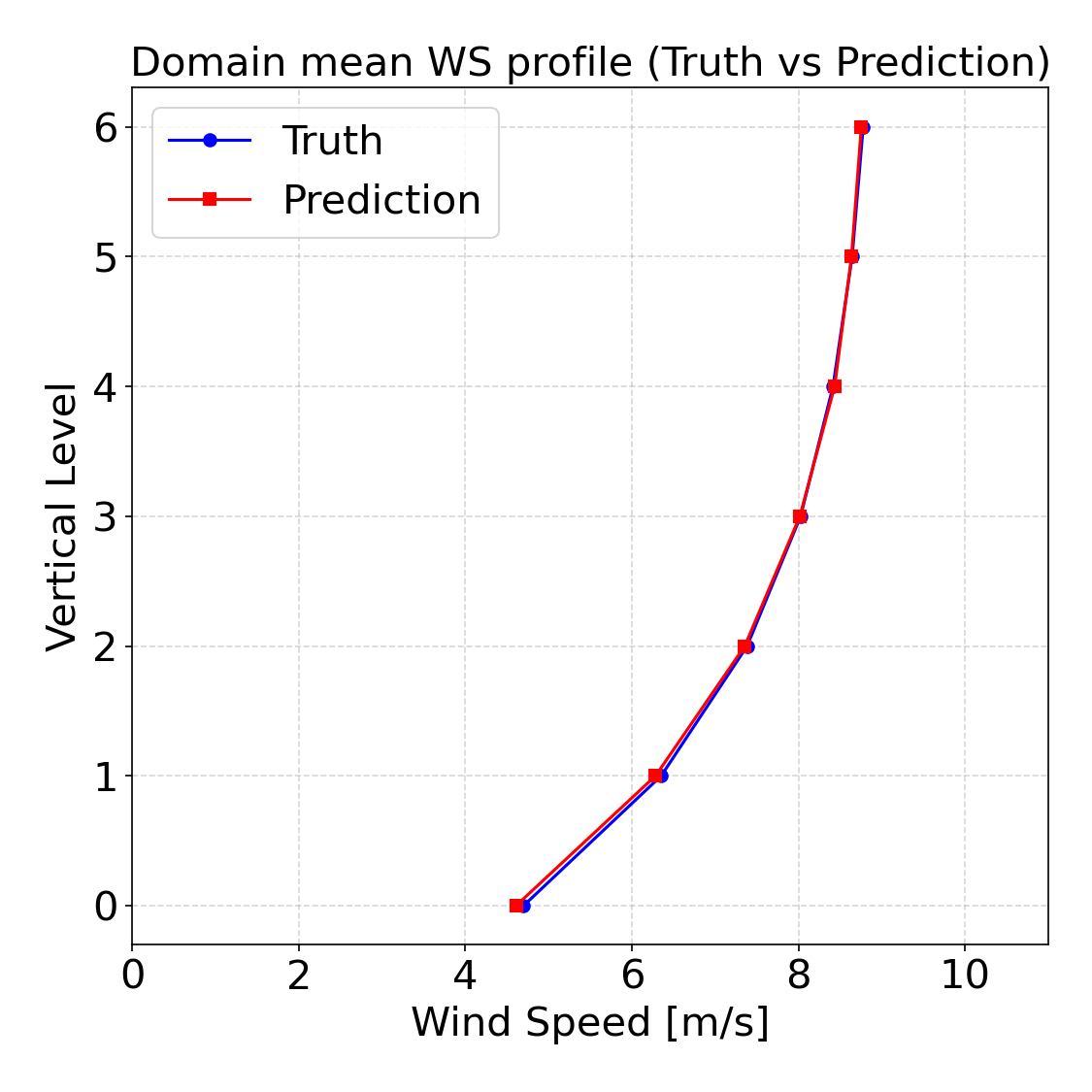}
        \label{fig:era5_rank_hist}
    }\hfill
    \subfloat[]{%
        \includegraphics[
            width=0.31\linewidth,
            trim=20 10 20 10,
            clip
        ]{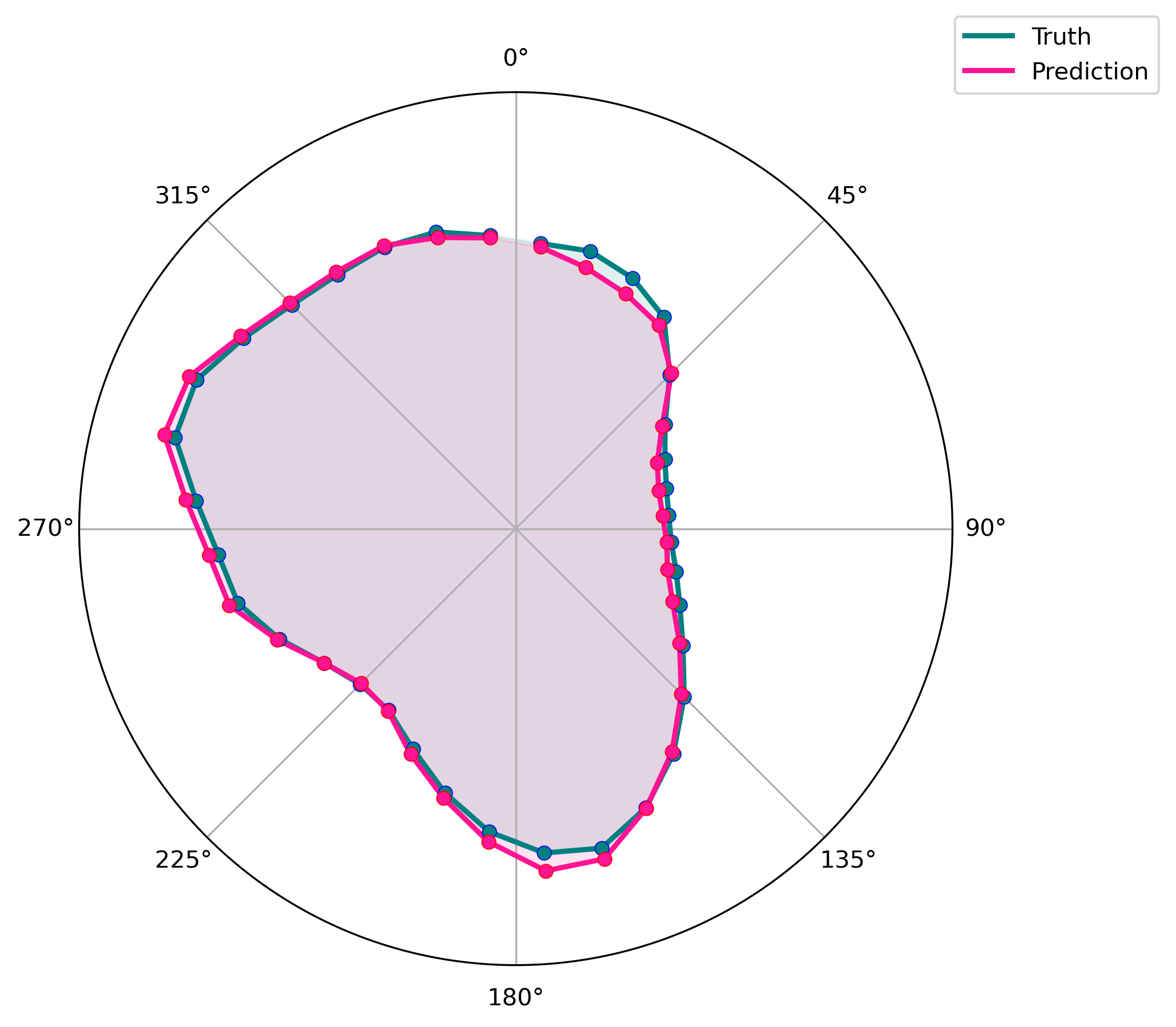}
        \label{fig:era5_polar_hist}
    }

    \par\medskip

    \subfloat[]{%
        \includegraphics[
            width=0.31\linewidth,
            trim=20 10 20 10,
            clip
        ]{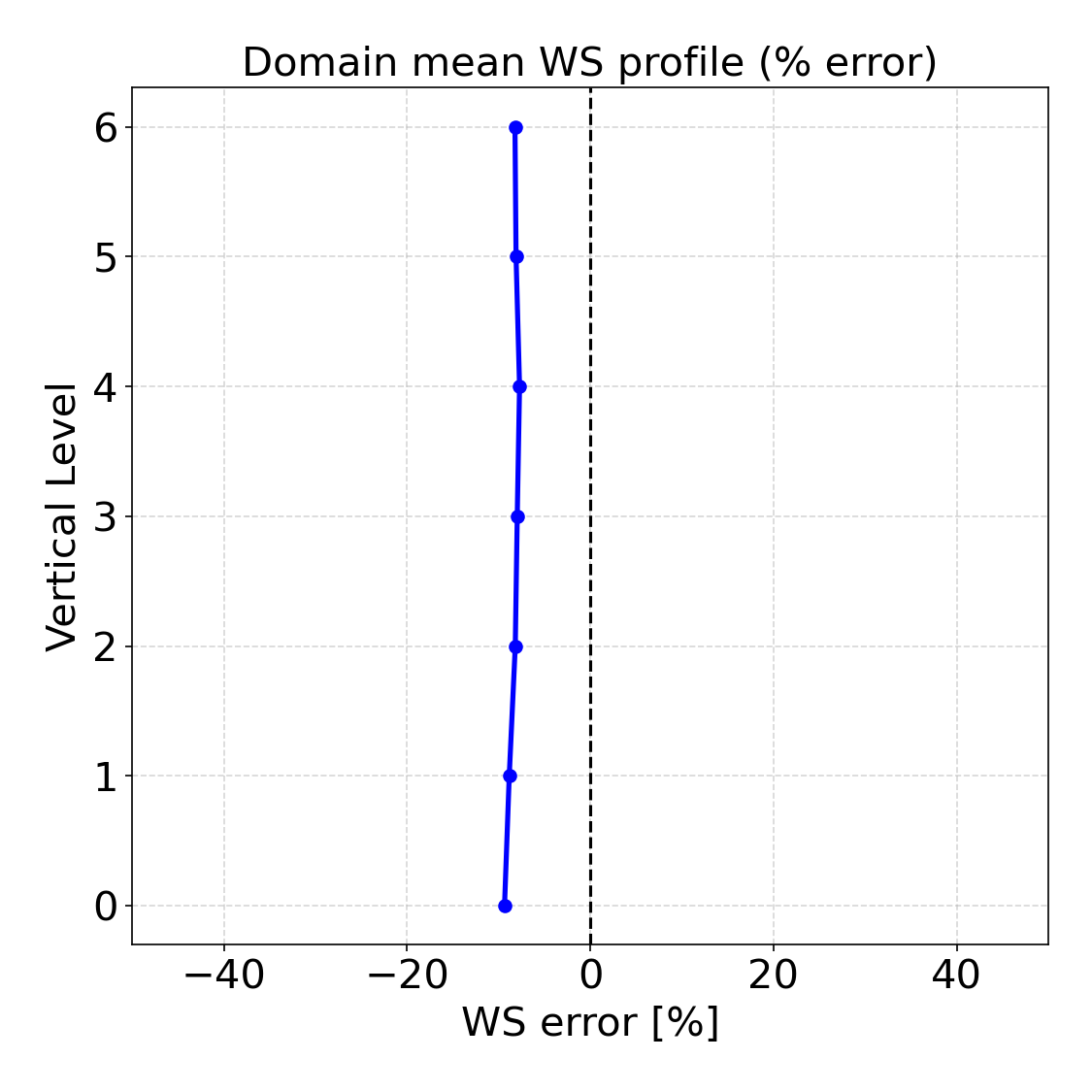}
        \label{fig:aurora_profile_WS}
    }\hfill
    \subfloat[]{%
        \includegraphics[
            width=0.31\linewidth,
            trim=20 10 20 10,
            clip
        ]{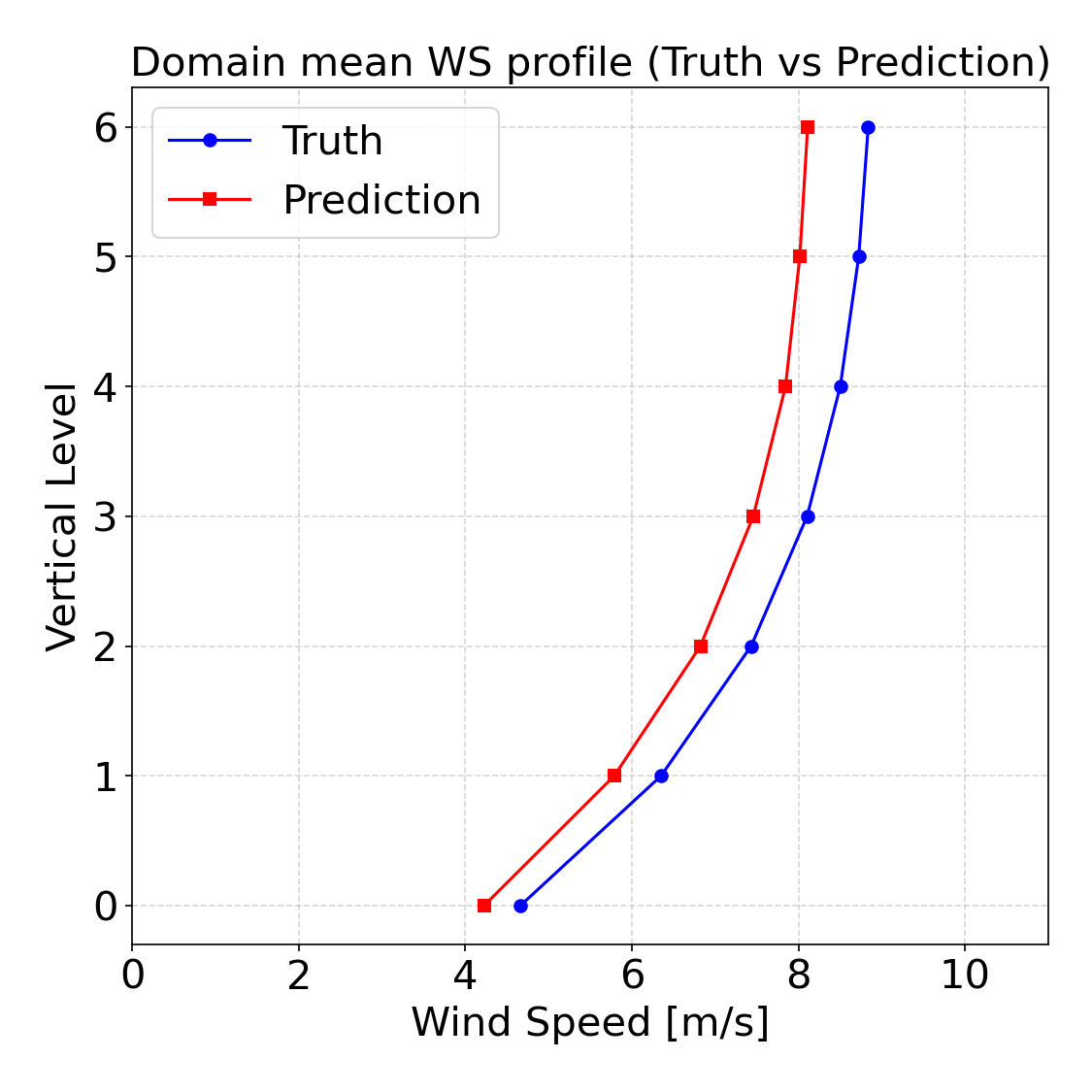}
        \label{fig:aurora_rank_hist}
    }\hfill
    \subfloat[]{%
        \includegraphics[
            width=0.31\linewidth,
            trim=20 10 20 10,
            clip
        ]{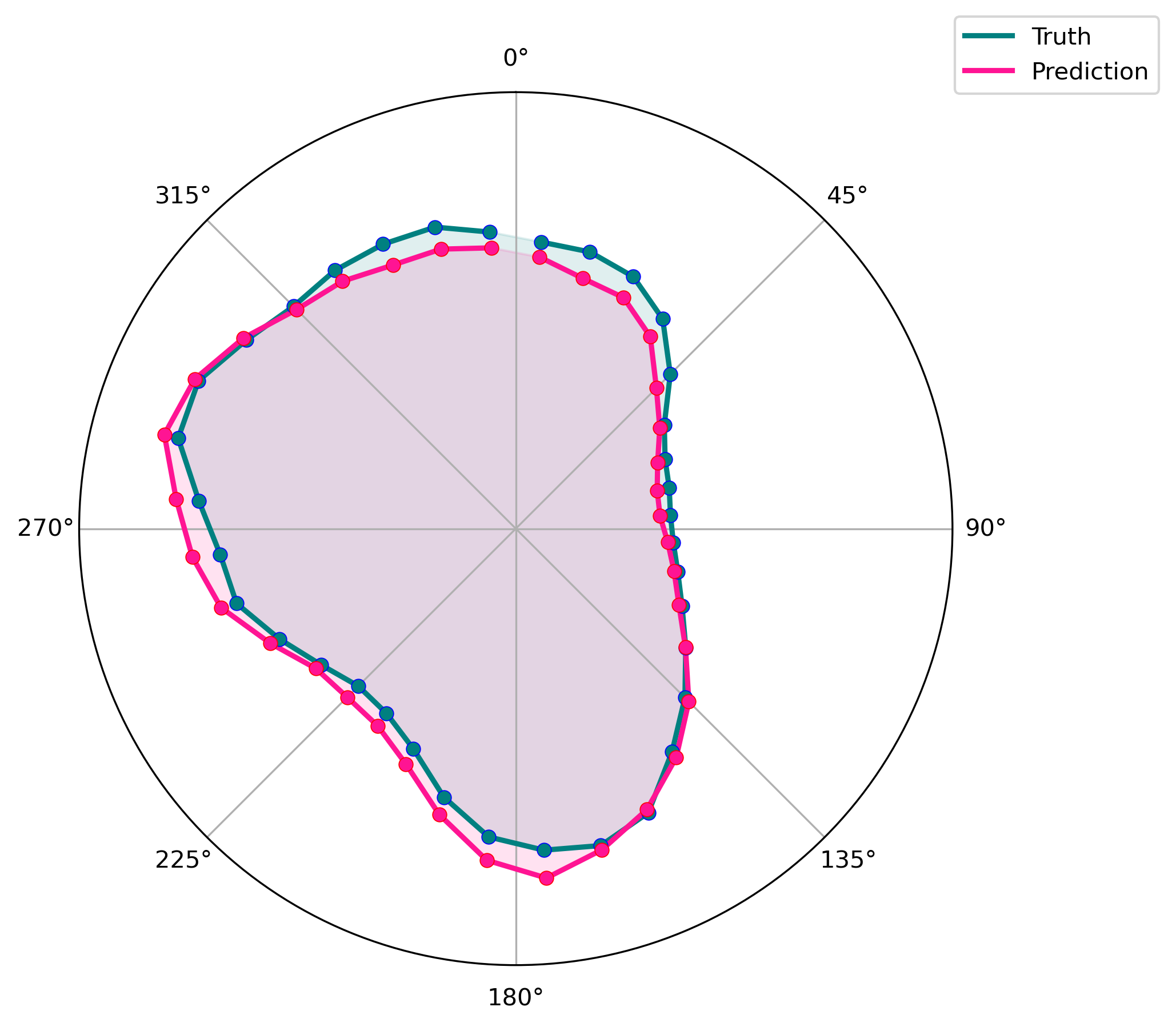}
        \label{fig:aurora_polar_hist}
    }

    \caption{Mean wind speed and direction statistics over the
    2014--2015 validation period, with ERA5 as input in the top row and
    Aurora as input in the bottom row. The columns show percent error in
    domain-mean wind speed by vertical level, vertical wind-speed profiles,
    and polar histograms of wind direction.}
    \label{fig:profile_combined}
\end{figure}

We evaluate the model’s ability to predict wind speed and direction at seven vertical levels not present in either the ERA5 input or the Aurora output. Fig.~\ref{fig:profile_combined}(a,b,d,e) compares the mean vertical profiles from the reference and predicted fields at seven reconstructed layers. The predicted profiles closely follow the reference across all levels, with average relative errors remaining below 3\% for the ERA5 input and 9.36\% for the Aurora input. 
The Aurora-based predictions show larger errors than the ERA5-based ones. This is expected because the model was trained using ERA5 data, while AURORA inputs are introduced only during inference. As a result, Aurora may introduce its own model biases, which can slightly shift the generated output from the WRF target fields.
We use Aurora in inference mode to provide the forecasting capability to the framework without additional training. Since the current work does not fine-tune Aurora for the target region or variables, any biases in Aurora's forecasts can propagate into the downscaling model. Fine-tuning Aurora for this regional forecasting setting may further improve the quality of the Aurora inputs and consequently the downstream predictions, and to further improve Aurora's predictions, it may require fine-tuning on our target domain; that direction is left for future work.
As illustrated in Fig.~\ref{fig:profile_combined} (a,d), the prediction bias is largest at the lowest level and decreases with height, approaching zero in the upper levels. We believe that this trend may be related to increased turbulence, variability, and sensitivity to surface processes near the boundary layer, while winds aloft tend to be more stable and therefore easier to predict.

\begin{figure}[!h]
  \centering
  \subfloat[ERA5]{\includegraphics[width=\linewidth]{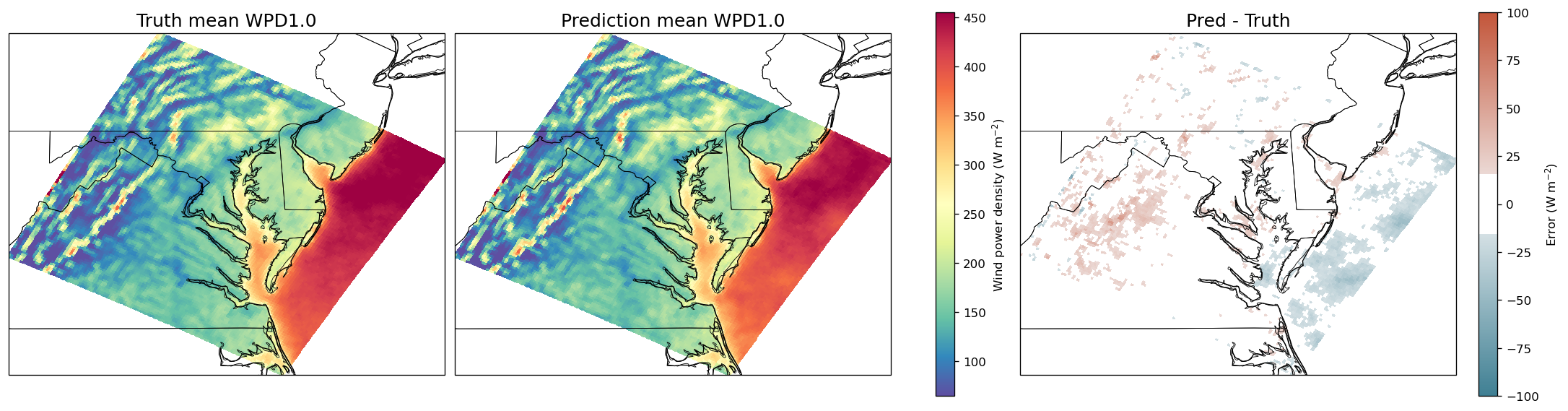}\label{fig:wpd_era5}}
  
  
  \subfloat[Aurora]{\includegraphics[width=\linewidth]{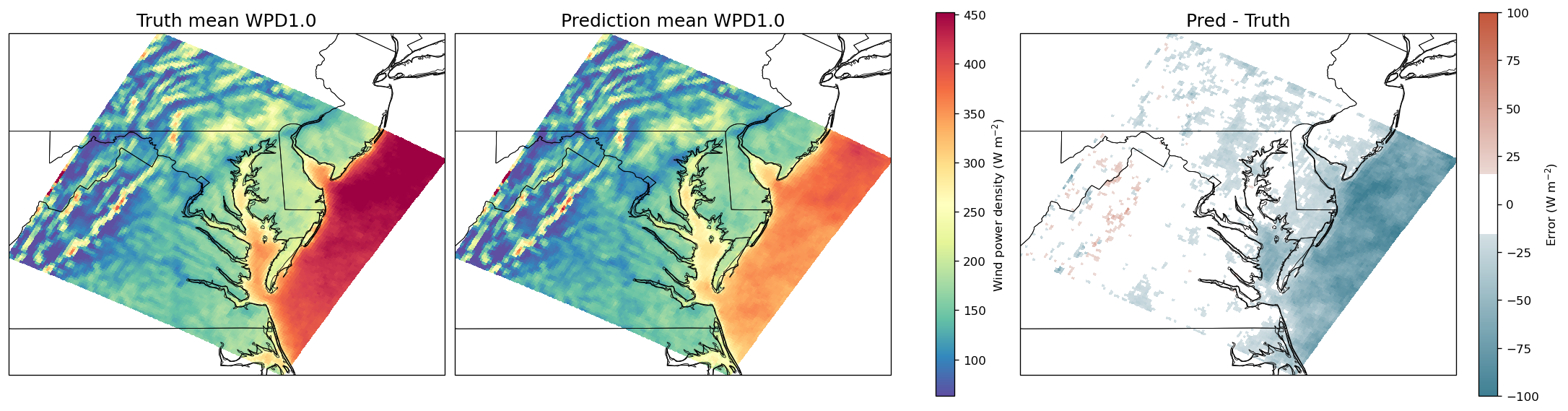}\label{fig:wpd_aurora}}
  
  \caption{Mean wind power density generated by CorrDiff through the downscaling process. The first column from left shows the reference WRF fields, the second column shows the CorrDiff predictions, and the third column shows the prediction bias ($Prediction - Truth$). Results are shown for (a) ERA5-driven inputs (top row) and (b) Aurora-driven inputs (bottom row).}
  \label{fig:wpd}
\end{figure}

In addition to the wind speed, Fig.~\ref{fig:wpd} shows the mean Wind Power Density of level one. The ground truth is computed from WRF outputs using WRF routines. The predicted results closely match the ground truth results. 

\subsection{Computational Efficiency}
Producing a full day of high-resolution WRF output typically requires a couple of hours of wall time on a multi-node CPU/GPU cluster. In contrast, Apeliotes performs a single forward pass per input timestamp on a single NVIDIA H100 GPU, so a full day of downsampled forecasting (four timestamps) costs 20 seconds when using one ensemble and can scale up to less than two minutes when using 32 ensembles. This yields a significant speedup, which emphasizes the model's capability to produce accurate and real-time downscaled atmospheric data.

\section{Conclusion and Future Work}
\label{sec:discussion}
This work explores downscaling from a global foundation model to a regional domain while directly generating new multi-level atmospheric fields.
We have demonstrated the competitive performance of \modelname, the coupled framework enables efficient high resolution weather predictions at 4\,km resolution. Furthermore the proposed framework can be trained to predict derived multi-level atmospheric fields, which are not readily provided by the input global model, including nonlinear and state-dependent variable WPD. This capability opens the door to a broad set of potential applications for \modelname.

A limitation of our current implementation is the training time required for the diffusion model. For the region covering Hampton Roads, VA, the model training requires multiple days on $8\times$ NVIDIA GPUs. As we further scale the covered target area to a larger region, more computing resources are necessary. The scaling issue is further compounded as we explore higher spatial resolution in order to provide forecasting at sub-kilometer resolutions.

In future work, we plan to further explore the generative capability of \modelname, to include additional atmospheric fields and application-oriented derived variables. We also plan to address the training efficiency of the diffusion component of \modelname. The efforts will enable \modelname~to be extended to sub-kilometer resolutions. Another important issue is the development of data assimilation solutions to incorporate sparse sensor data \cite{manshausen2025generative}. Lastly we plan to investigate the integration of other global foundation models, such as GraphCast~\cite{lam2023learning}, into our framework.

\clearpage
\bibliographystyle{unsrt}  
\bibliography{references}

\end{document}